\documentclass[11pt]{article}
\usepackage[preprint]{acl}
\usepackage{times}
\usepackage{latexsym}
\usepackage[T1]{fontenc}
\usepackage[utf8]{inputenc}
\usepackage{microtype}
\usepackage{inconsolata}
\usepackage{amsmath,amssymb}
\usepackage{graphicx}
\usepackage{placeins}
\usepackage{needspace}
\usepackage{booktabs,longtable,tabularx,array,calc}
\usepackage{newunicodechar}
\newunicodechar{β}{\ensuremath{\beta}}
\newunicodechar{⁻}{\textsuperscript{-}}
\newunicodechar{⁴}{\textsuperscript{4}}
\newunicodechar{⁵}{\textsuperscript{5}}
\newunicodechar{₁}{\textsubscript{1}}
\newunicodechar{₂}{\textsubscript{2}}
\newunicodechar{→}{\ensuremath{\to}}

\title{How to Use Parallel Data for Multilingual Pretraining}
\author{
  \textbf{Zhihong Huang} \\
  University of Macau \\
  \texttt{zh.huang5@siat.ac.cn}
  \And
  \textbf{Yao Lu} \\
  University College London \\
  \texttt{yao.lu@cs.ucl.ac.uk}
  \And
  \textbf{Derek F. Wong} \\
  University of Macau \\
  \texttt{derekfw@um.edu.mo}
  \AND
  \textbf{Hamid Alinejad-Rokny} \\
  UNSW Sydney \\
  \texttt{h.alinejad@unsw.edu.au}
  \And
  \textbf{Min Yang} \\
  SIAT, CAS \\
  \texttt{min.yang@siat.ac.cn}
}

\begin{document}
\maketitle

\begin{abstract}

High-quality human-translated parallel corpora can improve cross-lingual
transfer, but are typically too small for large-scale pretraining.
Machine-translated web corpora help overcome this limitation, making
parallel data available at pretraining scale. Standard multilingual
pretraining, however, samples source documents and their translations
independently, without explicitly using their parallel structure.
We investigate how to use the parallel structure of the data to improve
multilingual pretraining by concatenating each document with its
translation in the same training sequence. We compare this parallel
pretraining (PPT) approach with standard pretraining (SPT), training
models from scratch at matched token budgets. Surprisingly, despite the
benefits of parallel data reported in prior cross-lingual work, PPT alone
does not consistently improve language understanding over SPT.
Our further analysis suggests that although PPT builds a strong
cross-lingual foundation, as shown by its better use of shared information
across languages, the mismatch between parallel training contexts and
single-language task inputs limits its gains in general language
understanding. This finding motivates a two-stage approach following
the pretraining--fine-tuning paradigm: we first use PPT to build a
cross-lingual foundation, then continue with SPT to adapt it to standard
language modeling. In both high-resource and low-resource settings,
the two-stage approach consistently improves general language understanding and
cross-lingual task performance over standard pretraining alone at matched
training budgets.

\end{abstract}

\begin{figure*}[t]
\centering
\includegraphics[width=0.75\textwidth,keepaspectratio]{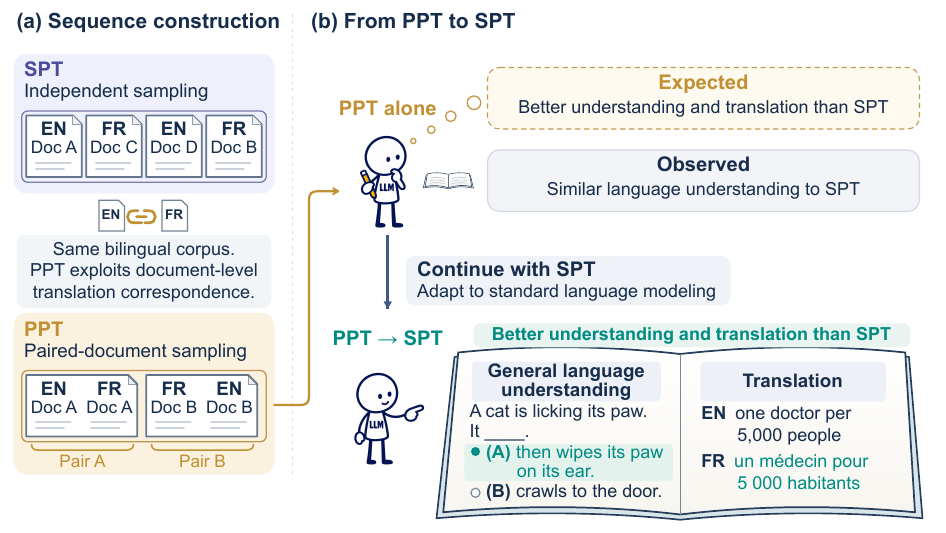}
\caption{Sequence construction and the motivation for PPT followed by SPT. (a) SPT samples documents independently; PPT joins translation pairs before sequence packing. Both use the same bilingual corpus. Letters identify content and EN/FR identify languages. (b) Continuing PPT with SPT improves general language understanding and translation over SPT at matched training budgets. The book pages illustrate these tasks using two HellaSwag continuations and an English--French translation excerpt.}\label{fig:training-design}
\end{figure*}
\section{Introduction}\label{introduction}

Multilingual parallel data has been widely used to improve cross-lingual language understanding~\citep{lample2019xlm,huang-etal-2019-unicoder,chi-etal-2021-infoxlm}. However, high-quality human-translated corpora are typically too small for large-scale multilingual pretraining, and scaling them is costly~\citep{nllb2024flores}. Machine translation makes it possible to translate high-quality English web corpora into a wide range of languages at pretraining scale~\citep{doshi2024translationese,wang2025transwebllm}. This reduces the resource gap by providing more high-quality training data for languages with lower-quality web data than English. Standard multilingual pretraining uses translated documents as an additional training source, treating original documents and their translations as independent examples and sampling them independently~\citep{wang2025transwebllm}.

However, treating these documents independently does not make direct use of their parallel structure. Previous work has used parallel data to improve cross-lingual language modeling and to adapt pretrained models to translation and other multilingual tasks~\citep{lample2019xlm,lee2022docmt5,kondo2024parallel,alves2024tower}. \citet{qorib2025jgp} study how to incorporate sentence-level parallel data\footnote{Their experiments use less than 5B tokens of parallel data.} into predominantly non-parallel pretraining. These studies show that parallel structure can improve cross-lingual capabilities even with relatively small amounts of parallel data~\citep{lin2025recipe}. With parallel documents available at pretraining scale~\citep{wang2025transwebllm,finetranslations2026}, it remains unclear how to use their parallel structure effectively for multilingual pretraining.

We investigate this question by introducing parallel pretraining (PPT), which concatenates each document with its translation in the same training sequence. We compare PPT with standard multilingual pretraining (SPT), which treats the original and translated documents as independent training examples (Figure~\ref{fig:training-design}(a)). To avoid uncertainty about what existing open-weight models have already learned from parallel data, we train both models from scratch. We use the same training setup and hyperparameters. The only difference is whether a document and its translation are treated as a single training sequence.

We pretrain 1.3B-parameter dense models with the Llama 2 architecture~\citep{touvron2023llama2} on TransWebEdu~\citep{wang2025transwebllm}. Our experiments cover English--French and English--Swahili, representing high-resource and low-resource settings, respectively. We train both SPT and PPT models for approximately 200B tokens and evaluate their performance on general language understanding and translation tasks.

Surprisingly, PPT alone does not consistently improve general language understanding over SPT. Further analysis shows that PPT models benefit more from cross-lingual context than SPT models. However, general language understanding tasks typically use single-language inputs, unlike the parallel sequences used in PPT. This context mismatch may limit how much these tasks benefit from PPT's cross-lingual capabilities.

This finding motivates a two-stage approach following the
pretraining--fine-tuning paradigm~\citep{howard-ruder-2018-universal}: PPT develops cross-lingual capabilities, and subsequent SPT adapts the model to standard language modeling (Figure~\ref{fig:training-design}(b)). We start the second stage from intermediate pretraining checkpoints and continue with SPT for 20B tokens. We also test SPT followed by PPT. We compare both two-stage approaches with SPT-only and PPT-only training
at the same total training budget. 
At a total training budget of 120B tokens, our experimental results show that PPT followed by SPT improves average general language understanding over SPT alone in both language settings. It also achieves the best translation performance among the four training strategies, with relative improvements in BLEU over SPT alone of 77\% on English--French and 45\% on English--Swahili.\footnote{BLEU increases from 19.25 to 34.03 on English--French and from 9.07 to 13.18 on English--Swahili.}

Our behavioral analyses show that continuing PPT with SPT improves monolingual language modeling. The resulting model also benefits more from cross-lingual context than a model trained with SPT alone.

To summarize, our contributions are as follows:
\begin{itemize}
\setlength{\itemsep}{2pt}
\setlength{\parsep}{0pt}
\item We investigate how to use parallel data at pretraining scale through
controlled comparisons of multilingual models trained from scratch.
\item Our proposed two-stage pretraining approach, which uses PPT followed
by SPT, shows improvements in both general language understanding and
translation over standard pretraining alone at matched training budgets.
\item Our behavioral analyses show that continued SPT improves monolingual
language modeling while retaining an advantage over SPT alone in using
cross-lingual context.
\end{itemize}

\section{Pretraining with Parallel Data}\label{pretraining-experiments}

\subsection{Independent and paired-document sequences}\label{data-and-sequence-construction}

Following \citet{wang2025transwebllm}, standard pretraining (SPT) samples
source and translated documents independently. Parallel pretraining (PPT)
concatenates each document with its translation in the same training
sequence (Figure~\ref{fig:training-design}(a)). Each document is preceded
by its language name, a colon, and a newline, giving the prefixes
``\texttt{English:\textbackslash n}'' and ``\texttt{French:\textbackslash n}''
in English--French.
We alternate which language comes first across document
pairs, preserving sentence order within each document.

Both methods train on the same bilingual corpus with autoregressive
next-token prediction. Both languages contribute to the loss, and
attention can span preceding documents within each 2,048-token window.
Pairing documents makes corresponding cross-lingual content available
when predicting the later document, alongside its preceding text in the
same language. Alternating document order provides this context in both
directions. Appendix~\ref{appendix-a-model-and-training-settings}
describes document matching, separators and training windows.

\subsection{From parallel to standard pretraining}\label{continued-pretraining}

PPT provides corresponding cross-lingual content during training, whereas
general language-understanding tasks typically use single-language inputs.
We therefore continue PPT with SPT to adapt the model to independently
sampled documents, denoting this approach by PPT→SPT
(Figure~\ref{fig:training-design}(b)). In this second stage, the model
continues learning from both languages without systematically receiving
the translation of the document it is predicting.

We compare PPT→SPT with SPT-only and PPT-only training at matched token
budgets. We also evaluate SPT→PPT to examine how the order of paired-document
and independent-document training affects language understanding,
translation, and use of cross-lingual context. Section~\ref{evaluation}
gives the training schedules and language settings.

\begin{figure*}[!t]
\centering
\includegraphics[width=0.95\textwidth,keepaspectratio]{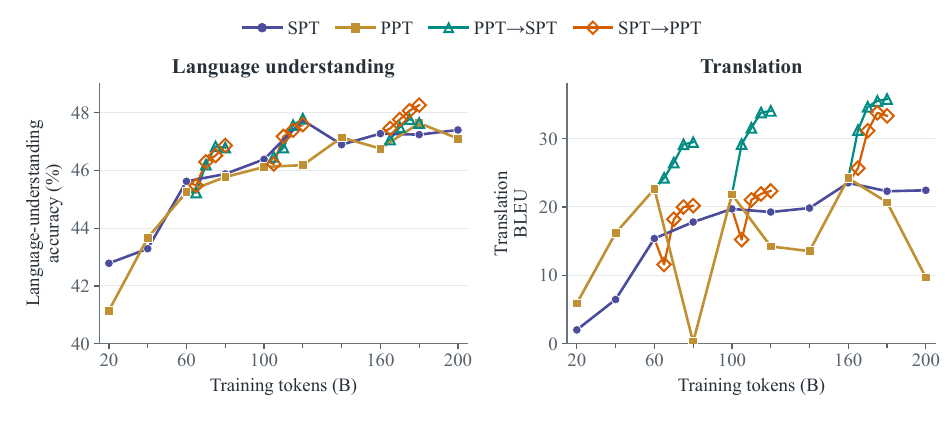}
\caption{English--French pretraining with SPT, PPT, and their staged combinations. SPT and PPT curves contain all ten checkpoints from 20B to 200B tokens. Each staged run branches from its initial method at 60B, 100B or 160B and trains for another 20B tokens, with evaluations every 5B tokens in both panels. Each branch is a separate training run. Accuracy averages six tasks and both languages equally. BLEU equally weights FLORES-200 and WMT14, both translation directions and the two prompt formats specified in Section~\ref{evaluation}.}\label{fig:training-trajectories}
\end{figure*}

\section{Experiments}\label{results}

\subsection{Experimental setup}\label{evaluation}

\textbf{Data and training.} We use English documents from TransWebEdu \citep{wang2025transwebllm}
and their corresponding French or Swahili translations, with approximately
150B available tokens per language. Each model is a 1.3B-parameter
decoder-only Transformer using the TransWebLLM architecture and tokenizer
\citep{wang2025transwebllm}. Within each language pair, the SPT and PPT
models train separately from the same initial weights for approximately
200B tokens, saving checkpoints every 20B.

To compare the two training orders at different stages, we continue
English--French models from 60B, 100B and 160B tokens with the other
sequence construction for another 20B tokens. These early-, middle-
and late-stage continuations use the same sampled TransWebEdu data
and end at 80B, 120B and 180B tokens, respectively.
English--Swahili uses the middle-stage comparison, from 100B to 120B.
Both orders use a new optimizer with learning-rate re-warming
\citep{gupta2023rewarm, ibrahim2024scalable} followed by cosine decay.
We compare the resulting models with SPT and PPT at the same total
training budget; these baselines retain their original schedule.
Appendix~\ref{appendix-a-model-and-training-settings} gives the model,
optimization and data-construction details.

\textbf{Evaluation.} We assess multilingual language understanding
with tasks in each language and cross-lingual generation with translation.
We evaluate language understanding with lm-evaluation-harness
\citep{eval-harness}. English--French
uses ARC-Challenge \citep{clark2018arc}, HellaSwag
\citep{zellers2019hellaswag}, TruthfulQA \citep{lin2022truthfulqa},
PAWS-X \citep{yang2019pawsx}, XNLI \citep{conneau2018xnli} and
XWinograd \citep{tikhonov2021xwinograd} in both languages.
English--Swahili uses XNLI, XStoryCloze \citep{lin2022xglm},
COPA/XCOPA \citep{roemmele-etal-2011-copa, ponti2020xcopa}
and Global-MMLU-Lite \citep{singh2024globalmmlu}.
Accuracy is averaged equally over tasks within
each language and then over the two languages.
Tasks use five sampled demonstrations, except English TruthfulQA MC1,
which uses six fixed examples without additional demonstrations.
French MC1 uses the same fixed examples plus five sampled demonstrations.
For each task, all models use the same demonstrations.

Translation uses FLORES-200 \citep{nllb2024flores} for both language pairs
and WMT14 \citep{bojar2014wmt14} for English--French, in both directions.
We compute corpus BLEU with SacreBLEU \citep{post2018sacrebleu},
weighting datasets, directions and two fixed prompting formats equally
within each language pair.
Appendix~\ref{appendix-b-evaluation-details} specifies the prompts and task protocols. Appendix~\ref{appendix-e-results-by-language-and-task}
reports scores for each language-understanding task and translation
condition at matched training budgets.

\subsection{English--French results}\label{english-french-results}

\subsubsection{SPT and PPT throughout pretraining}\label{complete-ppt-and-spt-training-runs}

Language-understanding accuracy generally rises with both SPT and PPT
(Figure~\ref{fig:training-trajectories}). Their bilingual averages differ
by at most 1.64 percentage points across checkpoints. PPT throughout
pretraining does not yield a sustained advantage over SPT in average
language understanding.

SPT rises from 2.03 to 22.43 mean BLEU, while PPT fluctuates more.
PPT leads during the first 60B
tokens, falls to 0.24 BLEU at 80B, and subsequently exceeds SPT
at 100B and 160B. At 200B, SPT exceeds PPT by 12.74
BLEU. The translation ranking thus changes during training.

\subsubsection{Combining PPT and SPT}\label{combining-ppt-and-spt}

\begin{table*}[!t]
\centering\small
\renewcommand{\arraystretch}{1.18}
\setlength{\tabcolsep}{3pt}
\begin{tabularx}{\textwidth}{@{}l*{8}{>{\raggedleft\arraybackslash}X}@{}}
\toprule
 & \multicolumn{4}{c}{Language understanding (\%)} & \multicolumn{4}{c}{Translation (BLEU)} \\
\cmidrule(lr){2-5}\cmidrule(l){6-9}
Tokens & SPT & PPT & PPT$\to$SPT & SPT$\to$PPT & SPT & PPT & PPT$\to$SPT & SPT$\to$PPT \\
\midrule
80B & 45.88 & 45.77 & 46.79 & \textbf{46.86} & 17.79 & 0.24 & \textbf{29.46} & 20.18 \\
120B & 47.74 & 46.18 & \textbf{47.79} & 47.59 & 19.25 & 14.23 & \textbf{34.03} & 22.34 \\
180B & 47.24 & 47.64 & 47.64 & \textbf{48.26} & 22.29 & 20.68 & \textbf{35.77} & 33.32 \\
\bottomrule
\end{tabularx}
\caption{English--French results at matched training-token budgets. Accuracy averages six tasks and both languages equally. BLEU averages both datasets, directions and prompt formats equally. Bold marks the highest value for each budget and metric. Staged models at different budgets are trained independently.}\label{tab:enfr-results}
\end{table*}

\textbf{Language understanding.} PPT→SPT improves accuracy over SPT
at all three matched budgets and matches or exceeds PPT
(Table~\ref{tab:enfr-results}). SPT→PPT exceeds both baselines
at 80B and 180B and scores 0.15 percentage points below SPT at 120B.
In both orders, accuracy exceeds the starting value after 10B tokens
and remains higher at 20B (Figure~\ref{fig:training-trajectories}).

Averaged equally over English and French, accuracy improves on commonsense
reasoning (HellaSwag) and grade-school science question answering (ARC-Challenge)
in both orders at all three stages relative to continuing the initial method.
PPT→SPT gains 1.83--3.67 percentage points on HellaSwag, and SPT→PPT
gains 1.67--2.08 points. These task gains contribute differently to
the overall average. At
180B, PPT→SPT and PPT both average 47.64\%. Relative to PPT,
PPT→SPT gains 3.67 percentage points on HellaSwag and loses 2.69
points on XWinograd. Table~\ref{tab:tasks-enfr}
gives the task-by-language results underlying these averages.

\textbf{Translation.} PPT→SPT increases BLEU at every 5B evaluation
within each continuation (Figure~\ref{fig:training-trajectories}). SPT→PPT exceeds its starting
score after 10B tokens at all three stages, following an initial drop
at the early and middle stages.
Both orders finish above SPT and PPT alone at every matched budget
(Table~\ref{tab:enfr-results}). PPT→SPT gains 11.68--14.78 BLEU over SPT, compared with
2.40--11.03 for SPT→PPT. Table~\ref{tab:enfr-translation} gives
the individual translation conditions.

\subsection{English--Swahili results}\label{low-resource-pretraining}

We extend the middle-stage comparison to English--Swahili to examine
whether PPT followed by SPT also benefits a low-resource language.
Across the 200B SPT and PPT runs, their bilingual averages differ by at most
1.44 percentage points, with neither method consistently ahead.
Table~\ref{tab:ensw-results} compares the two combinations with SPT and PPT alone
at 120B tokens, reporting each language as well as the bilingual mean.

\begin{table}[!t]
\centering\small
\setlength{\tabcolsep}{3pt}
\renewcommand{\arraystretch}{1.2}
\begin{tabular}{@{}lrrrr@{}}
\toprule
 & \multicolumn{3}{c}{Understanding (\%)} & BLEU \\
\cmidrule(lr){2-4}
Model & EN & SW & Mean & \\
\midrule
SPT & 53.08 & 45.69 & 49.38 & 9.07 \\
PPT & 51.12 & 45.02 & 48.07 & 12.28 \\
PPT$\to$SPT & 52.51 & \textbf{47.72} & \textbf{50.11} & \textbf{13.18} \\
SPT$\to$PPT & \textbf{54.24} & 45.98 & \textbf{50.11} & 1.54 \\
\bottomrule
\end{tabular}
\caption{English--Swahili results at 120B tokens. Understanding scores average four tasks per language; Mean weights English and Swahili equally. BLEU averages both directions and both prompt formats on FLORES-200 (Table~\ref{tab:ensw-translation}). Bold marks the highest score in each column.}\label{tab:ensw-results}
\end{table}

Both orders reach a bilingual average of 50.11\%, exceeding both
baselines at the same budget (Table~\ref{tab:ensw-results}). PPT→SPT achieves
the highest Swahili mean, with gains on all four Swahili tasks over
both baselines. SPT→PPT gives the highest English mean
and also improves the Swahili mean over both baselines.
Table~\ref{tab:ensw-tasks} provides the task-level results.

For translation, PPT→SPT improves over SPT from 9.07 to 13.18 BLEU
and also exceeds PPT's 12.28. The gain is concentrated in English-to-Swahili
translation (Table~\ref{tab:ensw-translation}). As in English--French, following PPT
with SPT improves both language understanding and translation over
standard pretraining. The reverse order reaches the same bilingual
understanding average as PPT→SPT and scores 1.54 BLEU in translation.

\begin{figure*}[!t]
\centering
\includegraphics[width=\textwidth,keepaspectratio]{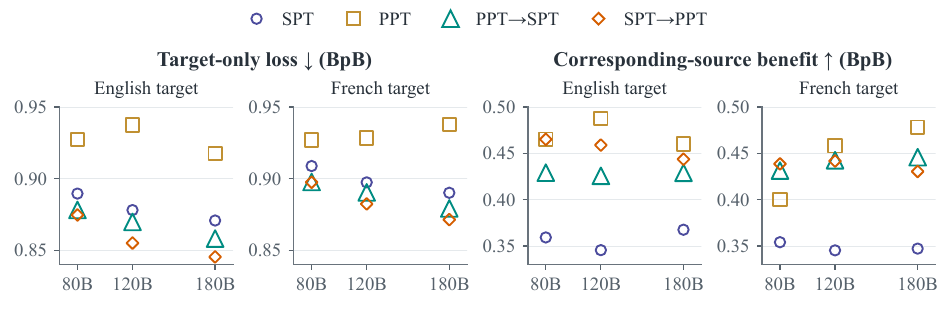}
\caption{Fixed-target prediction on the English--French WMT14 news test set. The left pair shows target-only loss, measuring monolingual prediction; the right pair shows corresponding-source benefit, the unrelated-minus-corresponding source loss. Each pair compares English and French targets on a common scale. Points average 173 document pairs in bits per byte; staged runs are independent. Table~\ref{tab:enfr-prediction-changes} gives differences from the corresponding SPT or PPT baseline at matched budgets, with 95\% document-bootstrap intervals.}\label{fig:document-prediction}
\end{figure*}

\section{Analysis of Prediction and Context Use}\label{prediction-abilities-after-a-format-switch}

PPT alone does not consistently improve general language understanding,
whereas PPT→SPT improves both understanding and translation.
We examine these results through monolingual language modeling and
cross-lingual context use, asking how they change when SPT follows PPT.
We then test the contribution of corresponding content and examine
answer changes on single-language tasks, comparing both training orders.

\subsection{Language modeling and cross-lingual context use}\label{independent-and-bilingual-context-prediction}

We score identical target continuations under three conditions to measure
monolingual language modeling and cross-lingual context use. Following
likelihood-based analyses in translation
\citep{fernandes-etal-2021-measuring, mohammed-niculae-2024-measuring},
we vary the preceding context while keeping the target fixed.

We use the English--French news translation test set from WMT14
\citep{bojar2014wmt14}. Its document boundaries and sentence alignment
allow us to vary bilingual context while keeping the target passage fixed.
In each direction, the first target-language sentence supplies the
prefix, and the next two sentences are scored. The prefix and
continuation are presented alone, after the corresponding three-sentence
source passage, or after a token-length-matched passage from another
document in the source language. Every model scores the same target
tokens in all three conditions using teacher forcing. The analysis covers
173 of the test set's 176 document pairs, whose first three aligned
sentences contain at most 600 tokens per language.

We measure \emph{monolingual language modeling} by \emph{target-only loss},
computed with the target-language prefix alone. The \emph{corresponding-source benefit}
is the loss after the other-document passage minus the loss after the
corresponding passage. It measures the advantage of corresponding
content when both inputs include an equally long foreign-language
prefix. Loss is reported in bits per UTF-8 byte (BpB), giving both
languages a common unit, and averaged equally over documents.
Figure~\ref{fig:document-prediction} compares models at each token budget.
We obtain 95\% confidence intervals from 2,000 document-bootstrap samples,
using the same resampled documents for each model and condition.
Appendix~\ref{appendix-c-document-prediction-methods} gives passage construction
and paired comparisons.

PPT alone obtains greater corresponding-source benefit than SPT but
higher target-only loss (Figure~\ref{fig:document-prediction}). It uses
shared information across languages more effectively, while predicting
single-language text less accurately. At every matched budget,
PPT→SPT has lower target-only loss than SPT and PPT, while retaining
greater corresponding-source benefit than SPT in both languages.
SPT→PPT also has lower target-only loss and greater
corresponding-source benefit than SPT.

To examine the second stage itself, we also compare each model with its
starting checkpoint. During the 20B of SPT following PPT, target-only
loss falls by 0.038--0.061 BpB across the three stages and two languages.
The corresponding-source benefit decreases, while the final model
retains a greater benefit than SPT at the same budget. In the reverse
order, target-only loss falls by 0.020--0.029 BpB and corresponding-source
benefit increases in all six comparisons. Both orders improve monolingual
prediction, while changing the benefit of bilingual context in different
directions. Appendix~\ref{paired-model-comparisons}
reports corresponding-source changes and matched-budget comparisons.

We apply the same analysis to the aligned English--Swahili health
documents in AfriDocMT
\citep{alabi-etal-2025-afridoc}. We score every noninitial sentence
under the three conditions across the complete collection of 61 documents.
Both staged models achieve lower target-only loss and greater
corresponding-source benefit than SPT in Swahili and English
at 120B (Table~\ref{tab:ensw-prediction-changes}). During SPT after PPT,
target-only loss falls and corresponding-source benefit decreases from
the PPT starting point, as in English--French. The final model thus
combines improved monolingual language modeling with greater
cross-lingual context benefit than SPT in the low-resource setting.

\subsection{Using corresponding content}\label{using-corresponding-content}

\begin{figure*}[!t]
\centering
\includegraphics[width=\textwidth,keepaspectratio]{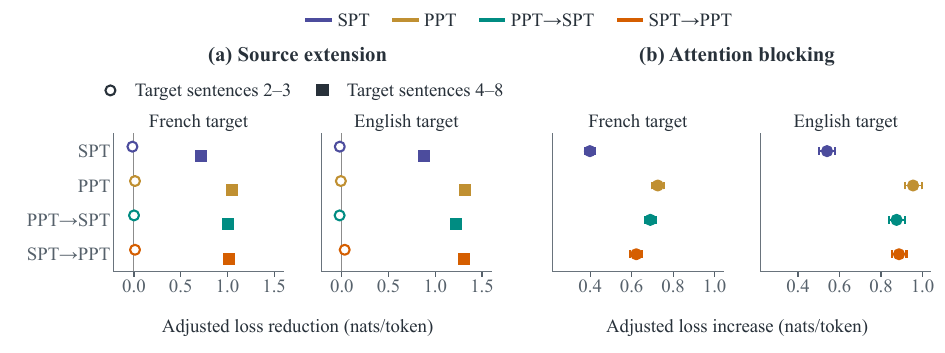}
\caption{Corresponding-content use on the English--French WMT14 news test set at 120B. Each pair compares French and English targets. (a) Loss reduction after extending source sentences 1--3 to 1--8, with the target fixed. Circles denote target sentences 2--3, whose source counterparts were already present; squares denote sentences 4--8, whose counterparts are added. (b) Loss increase on target sentences 4--8 after blocking attention to their source counterparts in the middle eight layers (9--16). Both effects subtract the same operation with an unrelated source. Points average 133 document pairs in nats per target token, with 95\% document-bootstrap intervals. Appendix~\ref{c.1-corresponding-content-extension-and-source-access} illustrates the operations and reports all three layer groups.}\label{fig:content-use}
\end{figure*}

We measure how adding source sentences and blocking attention to them
affect fixed-target prediction \citep{fernandes-etal-2021-measuring}.
Both analyses compare the four 120B models on all 133 WMT14 document
pairs whose first eight aligned sentences contain at most 900 tokens
per language.

For source extension, we fix target sentences 1--8, using the first as
a prefix and scoring sentences 2--8. Extending the source from sentences
1--3 to 1--8 supplies counterparts for target sentences 4--8; those for
sentences 2--3 were already available. We subtract the loss reduction
from extending a length-matched unrelated source. For each target region,
we average this adjusted reduction equally over documents, in nats per
target token.

Figure~\ref{fig:content-use}(a) shows that the loss reduction is concentrated
on sentences 4--8 in both target languages. For these sentences, PPT
and the staged models gain 1.00--1.32 nats per token, compared with
0.72--0.87 for SPT. For sentences 2--3, adjusted changes range from
$-$0.024 to 0.030 nats per token across all four models and both
target languages.

To test the contribution of source access, we adapt the attention-knockout
intervention of \citet{geva-etal-2023-dissecting}.
We retain the complete input and block attention from positions
predicting target sentences 4--8 to source sentences 4--8. The loss
increase is measured against the unmodified model on the same input,
then adjusted by subtracting the increase under an unrelated source.
The target-language prefix, preceding target tokens and source
sentences 1--3 remain accessible. To test this dependence at different
depths, we intervene separately in the early, middle and late thirds
of the model, each comprising eight layers (1--8, 9--16 and 17--24).
Figure~\ref{fig:content-use}(b) shows the middle group, where blocking
these connections has a larger effect on PPT and the staged models
than on SPT.
For French targets, the adjusted loss increase is 0.62--0.73 nats per
token for PPT and the staged models, compared with 0.40 for SPT.
For English targets, the corresponding values are 0.87--0.95 and 0.54.
Appendix~\ref{c.1-corresponding-content-extension-and-source-access}
provides the input diagrams and the separate interventions on earlier
and later layer groups.
PPT→SPT therefore retains greater sensitivity to corresponding content
than SPT on both measures; SPT→PPT also exceeds SPT.

The same interventions on all 61 AfriDocMT documents show this pattern
in English--Swahili. On Swahili target sentences 4--8, the adjusted
extension benefit is 0.62--0.69 nats per token for PPT and staged models,
compared with 0.37 for SPT. Both languages show greater extension benefits
and larger effects of blocking layers 9--16 for these models (Table~\ref{tab:ensw-content-use}).

\subsection{Commonsense reasoning and property knowledge}\label{judgments-without-parallel-context}

\begingroup\widowpenalty=10000
We next examine how answers change on single-language commonsense tasks.
We compare answers on English
and French HellaSwag \citep{zellers2019hellaswag} and paired
English--Swahili XStoryCloze stories \citep{lin2022xglm}. Both tasks ask for plausible
continuations of single-language inputs. Comparing the same questions
before and after training separates corrected errors from retained
correct answers. Corrected and lost answers are reported as shares of
all questions; their difference is the change in accuracy.\par\endgroup

From 100B to 120B, both training orders retain more correct answers
than continued use of the initial method on both tasks. On HellaSwag,
they also correct more errors in English and French
(Figure~\ref{fig:answer-changes}). On XStoryCloze, lost answers decrease
by 0.79 percentage points for PPT→SPT and 1.16 for SPT→PPT,
averaging English and Swahili. These reductions outweigh slightly fewer
corrections, yielding higher accuracy in each language
(Table~\ref{tab:ensw-answer-changes}).

The continuation tasks assess plausible outcomes in a given situation.
We also examine whether models associate properties with appropriate
concepts, using English COMPS-BASE minimal pairs
\citep{misra-etal-2023-comps} for the English--French models.
Each pair tests whether a property phrase receives higher probability
after an appropriate concept than after a distractor.
At 120B, PPT→SPT reaches 75.15\% accuracy, up from PPT's 67.96\%
and close to SPT's 75.52\%. SPT→PPT reaches 77.27\%. Both orders
improve over continued use of their initial method in all five property
categories (Table~\ref{tab:property-knowledge}).

\section{Discussion}\label{discussion}

SPT and PPT have similar average language understanding throughout
pretraining, while their English--French translation ranking varies.
Following PPT with SPT improves both
understanding and translation at matched budgets in both language settings.
The prediction analyses clarify this contrast. In English--French,
PPT alone benefits more from corresponding context than SPT but has
higher monolingual prediction loss. Following PPT with SPT reduces this loss below
both baselines at every tested English--French stage. Corresponding-context
benefit decreases from the PPT starting point but remains above SPT.
The English--Swahili PPT→SPT model similarly has lower monolingual loss
and greater context benefit than SPT at 120B. Together, these results support developing
cross-lingual context use through PPT and then adapting to independently
sampled documents through SPT.

PPT→SPT retains greater use of corresponding content than SPT.
Added source sentences help predict their target-language counterparts,
and blocking attention to them raises loss.
These observations extend analyses of related-document context
\citep{shi2024incontextpretraininglanguagemodeling, zhao-etal-2024-analysing}
to another language's expression of the same content.
On single-language tasks, greater retention of correct answers contributes
to commonsense-reasoning gains in both language pairs, and subsequent
SPT improves PPT's concept--property judgments. These task-level gains
accompany improved monolingual prediction.

Introducing PPT after SPT also improves monolingual prediction while
increasing corresponding-context benefit. This order improves
English--French translation and gives the highest language-understanding
average at 80B and 180B. In English--Swahili, both orders reach the same
understanding average. PPT→SPT gives the highest translation average
across all tested stages and both language pairs. Its gains in both
general language understanding and translation favor this order for
multilingual pretraining across high- and low-resource settings.

\section{Related Work}\label{related-work}

\textbf{Multilingual pretraining data.} FineTranslations translates
non-English FineWeb2 documents into English, producing over one trillion
translated tokens \citep{penedo2025fineweb2,finetranslations2026}.
Translation into other languages also supports pretraining
\citep{doshi2024translationese,wang2025transwebllm}. TransWebLLM uses
independently sampled translations of English web documents and achieves
competitive multilingual performance, including in Swahili
\citep{wang2025transwebllm}.

\Needspace{3\baselineskip}
\textbf{Pretraining with parallel corpora.} Translation language
modelling predicts masked tokens in concatenated parallel sentences
\citep{lample2019xlm}, while document-level encoder--decoder objectives
support document translation and cross-lingual generation
\citep{lee2022docmt5, lu-etal-2023-trip}. Recent decoder-only studies examine
sentence ordering and parallel-data composition
\citep{kondo2024parallel, lin2025recipe}. \citet{qorib2025jgp} obtain their
strongest translation results with a short parallel-sentence stage
after predominantly English pretraining. We compare both training orders
at document scale, including long parallel pretraining followed by a
short independent-sampling stage on the same bilingual corpus.

\textbf{Sequence composition and multilingual context.}
Random cross-document context can interfere with prediction
\citep{zhao-etal-2024-analysing}. Grouping related documents improves
language modeling and context use
\citep{zhao-etal-2024-analysing,shi2024incontextpretraininglanguagemodeling,staniszewski2025splice}.
For multilingual adaptation, \citet{wu2025crossic} concatenate
title-matched Wikipedia articles in different languages and compare
with using the same article pairs without concatenation.
Removing naturally occurring bilingual documents reduces translation,
and restoring their parallel
content recovers much of that performance \citep{shao2026mixed}.
Our analyses hold target content fixed to measure the predictive benefit
of corresponding cross-lingual context.

\section{Conclusion}\label{conclusion}

PPT uses document correspondence in machine-translated corpora to improve
cross-lingual context use, but does not consistently improve general
language understanding. Following PPT with SPT improves
language understanding and translation over standard pretraining at
matched budgets in
both high- and low-resource settings. Behavioral analyses show improved
monolingual language modeling while retaining an advantage over SPT in
using cross-lingual context. These findings support PPT followed by SPT
for multilingual pretraining using large-scale document-level parallel corpora.

\pagebreak
\clearpage
\section*{Limitations}\label{limitations}
\addcontentsline{toc}{section}{Limitations}

Our experiments use 1.3B-parameter models and two language pairs,
English--French and English--Swahili. Extending the study to larger
models and more language pairs would broaden our understanding of
how the two-stage use of parallel data benefits multilingual pretraining.

\bibliography{custom}

\appendix
\section*{Appendix}
The appendix provides training and evaluation details, prediction and
context-use analyses, and results by language and task.
\raggedbottom
\makeatletter
\setlength{\@dblfptop}{0pt}
\makeatother
\section{Model and Training Settings}\label{appendix-a-model-and-training-settings}

The main models use the TransWebLLM architecture and Llama-2 tokenizer
\citep{wang2025transwebllm, touvron2023llama2}. Training uses
Megatron-LM \citep{shoeybi2020megatron} with FlashAttention-2
\citep{dao2023flashattention2}. Table~\ref{tab:training-settings}
lists the model and optimization settings.

\begin{table}[!ht]
\centering\footnotesize
\setlength{\tabcolsep}{3pt}
\begin{tabularx}{\columnwidth}{@{}>{\raggedright\arraybackslash}X>{\raggedright\arraybackslash}X>{\raggedright\arraybackslash}X@{}}
\toprule
Setting & SPT or PPT alone & Staged continuation \\
\midrule
Parameters & 1.3B & Same architecture \\
Layers / attention heads & 24 / 16 & 24 / 16 \\
Hidden / feed-forward width & 2,048 / 5,504 & 2,048 / 5,504 \\
Position encoding / activation / normalization & RoPE / SwiGLU / RMSNorm & Same \\
Embeddings / dropout & Untied / none & Same \\
Optimizer & AdamW, β₁=0.9, β₂=0.95 & Same optimizer family \\
Weight decay / gradient clipping & 0.1 / 1.0 & 0.1 / 1.0 \\
Global batch / sequence length & 2,048 / 2,048 & 2,048 / 2,048 \\
Learning rate & 6×10⁻⁴, constant after warmup & 5×10⁻⁴ peak; 10⁻⁵ minimum \\
Warmup & 2,000 iterations & 480 iterations \\
Decay after warmup & None & Cosine \\
Iterations & 48,000 & 4,800 \\
Approximate consumed tokens & 200B & 20B additional \\
\bottomrule
\end{tabularx}
\caption{Model and training settings. Global batch size counts sequences; sequence length is in tokens.}\label{tab:training-settings}
\end{table}

SPT and PPT start from the same initial weights within each language pair. Base
checkpoints are saved every 4,800 iterations, approximately 20B tokens.
Staged endpoint comparisons follow 20B additional training tokens. English--French
comprises two base runs and six continuations; English--Swahili
comprises two base runs and two continuations.
Continued pretraining loads the starting model weights without restoring
the optimizer or random-number-generator state. Its learning-rate
schedule is specified above.

All training data come from TransWebEdu. The English--French
continuations use the same data sample at the early, middle and late
starting stages. SPT samples documents independently, and PPT pairs
source documents and translations by their shared UUIDs.
PPT separates the two prefixed documents with a newline.
English--Swahili uses ``\texttt{English:\textbackslash n}'' and
``\texttt{Swahili:\textbackslash n}''.
An end-of-document token follows each document in SPT and each document
pair in PPT. The tokenized text is divided into 2,048-token training
windows, which can cross document boundaries. Attention is causal
within each window, and position IDs continue across document boundaries.
End-of-document tokens contribute to the loss.

\section{Evaluation Details}\label{appendix-b-evaluation-details}

\begin{table*}[!t]
\centering\footnotesize
\setlength{\tabcolsep}{3pt}
\begin{tabularx}{\textwidth}{@{}>{\hsize=1.4\hsize\linewidth=\hsize\raggedright\arraybackslash}X>{\hsize=1.4\hsize\linewidth=\hsize\raggedright\arraybackslash}X>{\hsize=.8\hsize\linewidth=\hsize\raggedleft\arraybackslash}X>{\hsize=.9\hsize\linewidth=\hsize\raggedright\arraybackslash}X>{\hsize=.6\hsize\linewidth=\hsize\raggedleft\arraybackslash}X>{\hsize=.9\hsize\linewidth=\hsize\raggedright\arraybackslash}X@{}}
\toprule
Task & Language & Items & Split & \shortstack[r]{Sampled\\examples} & Metric \\
\midrule
\href{https://huggingface.co/datasets/allenai/ai2_arc}{ARC-Challenge} & English & 1,172 & test & 5 & acc\_norm \\
\href{https://huggingface.co/datasets/alexandrainst/m_arc}{mARC} & French & 1,169 & test & 5 & acc\_norm \\
\href{https://huggingface.co/datasets/Rowan/hellaswag}{HellaSwag} & English & 10,042 & validation & 5 & acc\_norm \\
\href{https://huggingface.co/datasets/alexandrainst/m_hellaswag}{mHellaSwag} & French & 9,338 & val & 5 & acc\_norm \\
\href{https://huggingface.co/datasets/truthfulqa/truthful_qa}{TruthfulQA MC1} & English & 817 & validation & 0 & acc \\
\href{https://huggingface.co/datasets/alexandrainst/m_truthfulqa}{mTruthfulQA MC1} & French & 787 & val & 5 & acc \\
\href{https://huggingface.co/datasets/paws-x}{PAWS-X} & English / French & 2,000 each & test & 5 & acc \\
\href{https://huggingface.co/datasets/facebook/xnli}{XNLI} & English / French / Swahili & 2,490 each & validation & 5 & acc \\
\href{https://huggingface.co/datasets/Muennighoff/xwinograd}{XWinograd} & English & 2,325 & test & 5 & acc \\
\href{https://huggingface.co/datasets/Muennighoff/xwinograd}{XWinograd} & French & 83 & test & 5 & acc \\
\href{https://huggingface.co/datasets/juletxara/xstory_cloze}{XStoryCloze} & English / Swahili & 1,511 each & eval & 5 & acc \\
\href{https://huggingface.co/datasets/aps/super_glue}{COPA} & English & 100 & validation & 5 & acc \\
\href{https://huggingface.co/datasets/cambridgeltl/xcopa}{XCOPA} & Swahili & 500 & test & 5 & acc \\
\href{https://huggingface.co/datasets/CohereLabs/Global-MMLU-Lite}{Global-MMLU-Lite} & English / Swahili & 400 each & test & 5 & acc \\
\bottomrule
\end{tabularx}
\caption{Language-understanding evaluations. Task names link to the dataset releases. Item counts are per language. Sampled examples exclude the fixed question--answer prefix in the TruthfulQA templates.}\label{tab:evaluation-tasks}

\end{table*}

\subsection{Language-understanding tasks}\label{b.1-language-understanding-tasks}

The task families are ARC \citep{clark2018arc}, HellaSwag
\citep{zellers2019hellaswag}, TruthfulQA \citep{lin2022truthfulqa},
PAWS-X \citep{yang2019pawsx}, XNLI \citep{conneau2018xnli}, XWinograd
\citep{tikhonov2021xwinograd}, XStoryCloze \citep{lin2022xglm},
COPA/XCOPA
\citep{roemmele-etal-2011-copa, ponti2020xcopa}, and
Global MMLU \citep{singh2024globalmmlu}. The released language variants
and evaluation subsets are listed in Table~\ref{tab:evaluation-tasks}.

For ARC-Challenge and HellaSwag, answers are selected by
character-length-normalized log likelihood, and accuracy is reported as
\texttt{acc\_norm}. Other tasks use \texttt{acc}.
The TruthfulQA task templates prepend six fixed English question--answer
pairs to each formatted question. English MC1 uses this fixed prompt
without additional sampled demonstrations; French MC1 adds five sampled
demonstrations. All other tasks use five sampled demonstrations.
These settings are shared across models, with demonstration-selection
seed 1234.

For each language pair, we average tasks within each language and then
average the two languages. Global-MMLU-Lite
combines its six subject groups by sample count before entering the
four-task average. Each benchmark receives equal weight.
French XWinograd contains 83 evaluated items, so a
one-item change corresponds to about 1.20 percentage points.

The French ARC and HellaSwag versions are the GPT-3.5-turbo translations
released through Okapi \citep{lai2023okapi}; mTruthfulQA also attributes
its translations to that project. XWinograd uses the release evaluated
by \citet{muennighoff-etal-2023-crosslingual}.

\subsection{Translation}\label{b.2-translation}

Translation evaluation uses FLORES-200 \citep{nllb2024flores} and WMT14
\citep{bojar2014wmt14}, with corpus BLEU computed using SacreBLEU
\citep{post2018sacrebleu}.

FLORES-200 uses all 1,012 devtest sentences, with demonstrations from
dev; WMT14 English--French uses all 3,003 test sentences, with
demonstrations from validation. Each condition uses five demonstrations,
greedy decoding and newline stopping. We report corpus BLEU
separately for each direction and prompt.

\noindent\begin{minipage}{\columnwidth}
\centering\footnotesize
\setlength{\tabcolsep}{3pt}
\begin{tabularx}{\columnwidth}{@{}l>{\raggedright\arraybackslash}X@{}}
\toprule
Prompt & English-to-French input \\
\midrule
Prompt A & \texttt{English:\ \{source\}\textbackslash{}nFrench:} \\
Prompt B & \texttt{English\ phrase:\ \{source\}\textbackslash{}nFrench\ phrase:} \\
\bottomrule
\end{tabularx}
\captionof{table}{English-to-French translation prompts. \texttt{\{source\}} is the source sentence; \texttt{\textbackslash n} denotes a newline.}\label{tab:translation-prompts}
\end{minipage}\par

Reverse-direction inputs exchange the language names. Swahili directions
replace the appropriate language label with \texttt{Swahili}. The
demonstrations include the reference translation after the target-language
label. The test prompt ends at that label, where the model generates
the translation.
The primary average gives equal weight to the two datasets,
two directions and Prompts A and B in English--French, and to two
directions and Prompts A and B in English--Swahili.

Each reported comparison keeps the prompt form fixed within a
condition. The English--Swahili four-model comparison uses
demonstration seed 1234 and identical FLORES inputs across models.
Appendix~\ref{appendix-e-results-by-language-and-task} reports the individual conditions underlying the averages.

\section{Language Modeling and Cross-Lingual Context Use}\label{appendix-c-document-prediction-methods}

\subsection{Prediction conditions and scoring}

Models score identical target continuations with teacher forcing under
three conditions, using the target prefix alone or preceding it with
a corresponding or unrelated source passage. Only continuation tokens
contribute to loss.

The English--French WMT14 news test set provides 176 document pairs,
with document boundaries and sentence alignment in the official SGML
files. We include all pairs whose first three aligned sentences contain
at most 600 tokens per language, yielding 173 pairs.
The first target sentence supplies the prefix and the next two
sentences are scored. The corresponding three-sentence source passage
supplies bilingual context. The unrelated condition uses an equally long
prefix from another news document. We exclude the target document and
its highest character-TF--IDF match before sampling the unrelated source.
All models and input conditions use the same prepared passages.

The English--Swahili analysis uses all 61 health documents released with
AfriDocMT \citep{alabi-etal-2025-afridoc}, comprising 1,982 aligned
sentences. We reconstruct document boundaries and score every noninitial
sentence once. The unrelated source comes from another document with the
same number of complete sentences, a similar relative position and the
nearest token length. The same target token IDs are scored in all three
conditions.

\begin{table*}[!t]
\centering\footnotesize
\setlength{\tabcolsep}{2.2pt}
\begin{tabular*}{\textwidth}{@{\extracolsep{\fill}}lllrrrr@{}}
\toprule
Training order & Target language & Training tokens & \shortstack{Target-only loss\\reduction} & \multicolumn{3}{c}{Corresponding-source benefit} \\
\cmidrule(lr){5-7}
 & & & [95\% CI] & Change [95\% CI] & \shortstack{SPT/PPT\\alone} & \shortstack{Combined\\training} \\
\midrule
SPT$\to$PPT & FR & 80B  & 0.011 [0.008, 0.015] & 0.085 [0.077, 0.091] & 0.354 & 0.439 \\
SPT$\to$PPT & FR & 120B & 0.015 [0.011, 0.019] & 0.096 [0.088, 0.104] & 0.345 & 0.442 \\
SPT$\to$PPT & FR & 180B & 0.019 [0.015, 0.023] & 0.083 [0.076, 0.090] & 0.347 & 0.431 \\
SPT$\to$PPT & EN & 80B  & 0.015 [0.011, 0.019] & 0.106 [0.096, 0.117] & 0.359 & 0.465 \\
SPT$\to$PPT & EN & 120B & 0.023 [0.019, 0.027] & 0.113 [0.105, 0.122] & 0.346 & 0.459 \\
SPT$\to$PPT & EN & 180B & 0.025 [0.022, 0.029] & 0.076 [0.065, 0.088] & 0.368 & 0.444 \\
PPT$\to$SPT & FR & 80B  & 0.029 [0.025, 0.034] & 0.031 [0.025, 0.038] & 0.400 & 0.432 \\
PPT$\to$SPT & FR & 120B & 0.038 [0.025, 0.056] & $-$0.016 [$-$0.046, 0.019] & 0.458 & 0.443 \\
PPT$\to$SPT & FR & 180B & 0.059 [0.044, 0.078] & $-$0.032 [$-$0.040, $-$0.025] & 0.478 & 0.446 \\
PPT$\to$SPT & EN & 80B  & 0.049 [0.044, 0.055] & $-$0.036 [$-$0.043, $-$0.029] & 0.465 & 0.429 \\
PPT$\to$SPT & EN & 120B & 0.068 [0.047, 0.095] & $-$0.062 [$-$0.093, $-$0.031] & 0.488 & 0.426 \\
PPT$\to$SPT & EN & 180B & 0.059 [0.053, 0.066] & $-$0.031 [$-$0.054, $-$0.004] & 0.460 & 0.429 \\
\bottomrule
\end{tabular*}
\captionof{table}{Prediction comparisons on 173 English--French WMT14 documents at matched training budgets. Baselines are SPT for SPT$\to$PPT and PPT for PPT$\to$SPT. Positive differences denote lower target-only loss or a larger corresponding-source benefit. The last two columns give the baseline and combined-training benefit values. All values are in bits per byte; intervals are 95\% document-bootstrap intervals.}\label{tab:enfr-prediction-changes}
\end{table*}

\begin{samepage}
For a scored continuation \(y\) and context \(c\), we compute
\[
\operatorname{BpB}(y\mid c)=
\frac{-\sum_{t=1}^{|y|}\log p(y_t\mid c,y_{<t})}
{\ln(2)\,\operatorname{bytes}(y)}.
\]
\end{samepage}
Lower scores indicate better prediction. Losses and byte counts are
summed within each document before documents are averaged equally.
The corresponding-source benefit is the unrelated-source score minus
the corresponding-source score for the same target. Paired differences
compare models on the same documents. We compute 95\% intervals from
2,000 document-bootstrap samples. All conditions and models in a
comparison use the same resampled document indices.

\subsection{Paired model comparisons}\label{paired-model-comparisons}

Second-stage changes compare each endpoint with its starting model,
at 60B, 100B or 160B for English--French and 100B for English--Swahili.
On the 173 WMT14 documents, source benefit decreases by 0.005--0.051 BpB during
PPT→SPT and increases by 0.076--0.134 BpB during SPT→PPT.

Table~\ref{tab:enfr-prediction-changes} compares each English--French staged model
with continued use of its initial construction at the same token budget.
Target-only loss reduction is baseline loss minus staged loss;
corresponding-source benefit change is staged benefit minus baseline
benefit. Table~\ref{tab:ensw-prediction-changes} reports the same two differences
for English--Swahili, using SPT at 120B as the common reference for both
staged models.

\noindent\begin{minipage}{\columnwidth}
\centering\footnotesize
\setlength{\tabcolsep}{3pt}
\begin{tabular}{@{}llrr@{}}
\toprule
Model & Target & \shortstack{Target-only loss\\reduction} & \shortstack{Corresponding-\\source benefit\\change} \\
\midrule
SPT$\to$PPT 120B & SW & 0.010 & 0.151 \\
SPT$\to$PPT 120B & EN & 0.015 & 0.111 \\
PPT$\to$SPT 120B & SW & 0.007 & 0.119 \\
PPT$\to$SPT 120B & EN & 0.008 & 0.091 \\
\bottomrule
\end{tabular}
\captionof{table}{English--Swahili prediction differences from SPT at 120B on 61 AfriDocMT health documents, weighted equally, in bits per byte. Positive values denote lower target-only loss or a larger corresponding-source benefit.}\label{tab:ensw-prediction-changes}
\end{minipage}\par

\subsection{Source extension and attention blocking}\label{c.1-corresponding-content-extension-and-source-access}

\begin{figure}[!ht]
\centering
\includegraphics[width=\columnwidth]{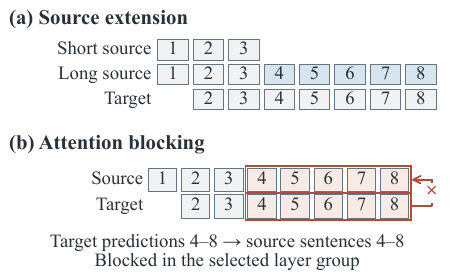}
\caption{Input and attention interventions for Figure~\ref{fig:content-use}. Numbered boxes denote aligned sentences. (a) The target is fixed while source sentences 4--8 are added in blue. (b) The complete input remains present; the crossed connection denotes blocked attention from target prediction positions to source sentences 4--8. Both operations are repeated with an unrelated source.}\label{fig:content-intervention-design}
\end{figure}

Figure~\ref{fig:content-intervention-design} illustrates the two operations.
Source extension uses WMT14 document pairs whose first eight aligned
sentences contain at most 900 tokens per language (133 pairs). The target prefix
and sentences 2--8 remain fixed. Extending the source from sentences
1--3 to 1--8 adds corresponding information for target sentences 4--8.
We also extend an unrelated source of matched length. Its short passage
is a prefix of the long passage from the same other document.
Figure~\ref{fig:content-use} reports
the reduction in mean target-token cross-entropy produced by the
corresponding extension minus the reduction produced by the unrelated
extension. The contrast is computed within each document before
averaging.

\begin{figure*}[!t]
\centering
\includegraphics[width=\textwidth,keepaspectratio]{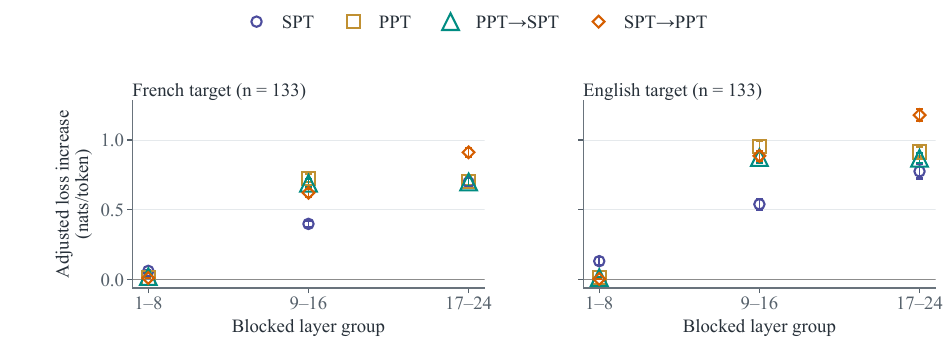}
\caption{Loss increase after blocking attention from positions predicting target sentences 4--8 to source sentences 4--8 in each layer group separately. Points average 133 English--French WMT14 documents at 120B, in nats per target token, with 95\% document-bootstrap intervals. The effect of blocking an unrelated source is subtracted. Both target languages share a scale. Layers 9--16 also appear in Figure~\ref{fig:content-use}(b).}\label{fig:attention-layers}
\end{figure*}

For the attention intervention, we adapt attention knockout
\citep{geva-etal-2023-dissecting} to the source and target sentence spans.
The complete source remains in the input. We block attention from positions predicting target sentences
4--8 to source sentences 4--8 in three equal eight-layer groups
(1--8, 9--16 and 17--24). Each group is blocked separately, comparing
different depths with the same intervention width. We measure the loss
increase from the unmodified model and subtract the increase under an
unrelated source.

Figure~\ref{fig:attention-layers} compares the three layer groups.
Blocking layers 9--16 or 17--24 produces larger adjusted loss increases
for PPT and both combined training orders than for SPT in both languages.
The effects of blocking layers 1--8 are smaller.

\subsection{Corresponding-content use in English--Swahili}

The same interventions use all 61 AfriDocMT health documents in both
directions. Source sentences 1--3 or 1--8 precede the first target
sentence; target sentences 2--8 are scored. Unrelated passages from
another document match the source's token length at each extent.
The target and unrelated document remain fixed within each comparison.

Table~\ref{tab:ensw-content-use} compares the four 120B models within each
intervention. Source extension separates target sentences 2--3 and
4--8; attention blocking compares the three layer groups. Both languages
show larger extension benefits and larger effects of blocking layers
9--16 for PPT and both combined training orders than for SPT.

\begin{table}[!ht]
\centering\footnotesize
\setlength{\tabcolsep}{2pt}
\begin{tabular*}{\columnwidth}{@{\extracolsep{\fill}}lrrrr@{}}
\toprule
Intervention & SPT & PPT & PPT$\to$SPT & SPT$\to$PPT \\
\midrule
\multicolumn{5}{@{}l}{Swahili target, source extension} \\
\shortstack[l]{Target sentences\\2--3} & $-$0.071 & $-$0.002 & $-$0.022 & $-$0.009 \\
\shortstack[l]{Target sentences\\4--8} & 0.372 & 0.686 & 0.622 & 0.671 \\
\multicolumn{5}{@{}l}{Swahili target, attention blocking} \\
Layers 1--8 & 0.006 & 0.030 & 0.020 & 0.003 \\
Layers 9--16 & 0.273 & 0.604 & 0.547 & 0.533 \\
Layers 17--24 & 0.308 & 0.249 & 0.231 & 0.500 \\
\midrule
\multicolumn{5}{@{}l}{English target, source extension} \\
\shortstack[l]{Target sentences\\2--3} & $-$0.086 & 0.016 & 0.003 & $-$0.002 \\
\shortstack[l]{Target sentences\\4--8} & 0.686 & 1.186 & 1.010 & 1.104 \\
\multicolumn{5}{@{}l}{English target, attention blocking} \\
Layers 1--8 & 0.408 & 0.024 & 0.039 & 0.017 \\
Layers 9--16 & 0.520 & 1.057 & 0.893 & 0.913 \\
Layers 17--24 & 0.618 & 0.439 & 0.385 & 0.883 \\
\bottomrule
\end{tabular*}
\caption{Corresponding-content interventions on 61 English--Swahili AfriDocMT documents at 120B. Source extension reports loss reduction for target sentences whose source counterparts were already present (2--3) or added (4--8). Attention blocking reports loss increase when prediction positions for target sentences 4--8 cannot attend to source sentences 4--8. Both effects subtract the unrelated-source effect and average documents equally, in nats per target token.}\label{tab:ensw-content-use}
\end{table}

\section{Judgments on Monolingual Inputs}\label{appendix-d-judgments-without-parallel-context}

Models share the items, candidates and scoring procedure within each
analysis. HellaSwag and XStoryCloze use five demonstrations in the
evaluated language. Answer changes compare the same questions before
and after training.

\subsection{HellaSwag answer changes}\label{d.1-hellaswag-across-english-and-french}

\begin{figure}[!ht]
\centering
\includegraphics[width=\columnwidth,keepaspectratio]{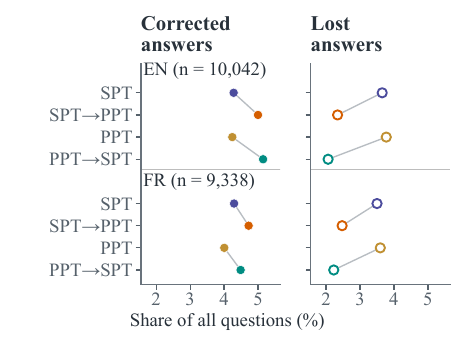}
\caption{Answer changes from 100B to 120B, as percentages of all 10,042 English HellaSwag or 9,338 French mHellaSwag questions. Corrected minus lost gives the accuracy change. Connected pairs share a 100B starting model.}\label{fig:answer-changes}
\end{figure}

Figure~\ref{fig:answer-changes} reports corrected and lost answers from 100B to 120B as
shares of all questions; their difference is the accuracy change.
We rescore all six models on the same questions to obtain item-level
predictions for this analysis. Table~\ref{tab:tasks-enfr}
reports the original benchmark scores for the four 120B models.

\subsection{English--Swahili answer changes}\label{d.2-englishswahili-answer-changes}

XStoryCloze \citep{lin2022xglm} provides aligned English--Swahili
stories, each with four context sentences and two candidate endings.
We evaluate the full set of 1,511 stories. With five-shot inputs in each language, the model
selects the ending with greater summed token log probability.
Answer-change results use a common item-level scoring run across
the six models; Appendix~\ref{appendix-e-results-by-language-and-task}
reports the original benchmark scores.

Table~\ref{tab:ensw-answer-changes} decomposes answer changes from each 100B starting model,
using all 1,511 stories per language as the denominator.
Corrected answers change from incorrect to correct; lost answers change
from correct to incorrect; retained answers remain correct.
Corrected plus retained correct gives the 120B accuracy in this scoring run.
\begin{table}[!ht]
\centering\footnotesize
\setlength{\tabcolsep}{2pt}
\begin{tabular*}{\columnwidth}{@{\extracolsep{\fill}}llrrr@{}}
\toprule
Language & 120B model & \shortstack{Corrected\\answers} & \shortstack{Lost\\answers} & \shortstack{Retained\\correct} \\
\midrule
\multicolumn{5}{@{}l}{Starting from SPT at 100B} \\
EN & SPT & 3.51 & 3.90 & 60.62 \\
 & SPT$\to$PPT & 3.38 & 2.45 & 62.08 \\
SW & SPT & 3.04 & 3.11 & 52.81 \\
 & SPT$\to$PPT & 2.65 & 2.25 & 53.67 \\
Mean & SPT & 3.28 & 3.51 & 56.72 \\
 & SPT$\to$PPT & 3.01 & 2.35 & 57.88 \\
\addlinespace
\multicolumn{5}{@{}l}{Starting from PPT at 100B} \\
EN & PPT & 4.10 & 3.64 & 59.50 \\
 & PPT$\to$SPT & 4.04 & 2.65 & 60.49 \\
SW & PPT & 2.65 & 2.58 & 53.21 \\
 & PPT$\to$SPT & 2.58 & 1.99 & 53.81 \\
Mean & PPT & 3.38 & 3.11 & 56.35 \\
 & PPT$\to$SPT & 3.31 & 2.32 & 57.15 \\
\bottomrule
\end{tabular*}
\caption{XStoryCloze answer changes on 1,511 paired English--Swahili stories. Adjacent rows compare continued and switched training from the same 100B model. Values are percentages of all questions; Mean averages the two languages equally.}\label{tab:ensw-answer-changes}
\end{table}

\subsection{Concept--property judgments}\label{d.3-concept-property-judgments}

COMPS-BASE \citep{misra-etal-2023-comps} uses English minimal pairs.
A comparison is correct when the property phrase receives higher
probability after the appropriate concept than after the distractor.
At 120B, accuracy over all 49,340 comparisons is 77.27\% for
SPT→PPT versus 75.52\% for SPT, and 75.15\% for PPT→SPT versus
67.96\% for PPT. Table~\ref{tab:property-knowledge} reports these
comparisons within each of five property categories, together with
the number of concept--property groups in each.
\begin{table}[!ht]
\centering\footnotesize
\setlength{\tabcolsep}{2pt}
\begin{tabularx}{\columnwidth}{@{}>{\hsize=2.05\hsize\linewidth=\hsize\raggedright\arraybackslash}X*{5}{>{\hsize=.79\hsize\linewidth=\hsize\raggedleft\arraybackslash}X}@{}}
\toprule
Model & Func. & Visual & Other & Encycl. & Taxon. \\
Groups, $n$ & 3,399 & 4,072 & 767 & 3,240 & 857 \\
\midrule
SPT 120B & 77.76 & 73.37 & 70.24 & 74.28 & 86.29 \\
SPT→PPT 120B & 79.76 & 74.83 & 71.77 & 76.38 & 87.25 \\
\midrule
PPT 120B & 70.40 & 67.15 & 60.98 & 66.10 & 75.32 \\
PPT→SPT 120B & 77.90 & 73.37 & 70.53 & 73.46 & 83.17 \\
\bottomrule
\end{tabularx}
\caption{COMPS-BASE accuracy (\%) on English minimal pairs for the 120B English--French models. Columns cover functional, visual perceptual, other perceptual, encyclopedic and taxonomic properties. The Groups row counts concept--property groups; each contributes four negative-concept comparisons.}\label{tab:property-knowledge}
\end{table}

\clearpage
\twocolumn[{
\begin{minipage}{\columnwidth}
\section{Detailed Benchmark Results}\label{appendix-e-results-by-language-and-task}

Tables~\ref{tab:tasks-enfr}--\ref{tab:ensw-translation} give the task and
translation scores underlying Tables~\ref{tab:enfr-results} and~\ref{tab:ensw-results}.
\end{minipage}
\par\vspace{\textfloatsep}

\begin{minipage}{\textwidth}
\begingroup
\centering\footnotesize
\setlength{\tabcolsep}{.5pt}
\begin{tabular*}{\textwidth}{@{\extracolsep{\fill}}lrrrr@{\hspace{6pt}}rrrr@{\hspace{6pt}}rrrr@{}}
\toprule
 & \multicolumn{4}{c}{80B} & \multicolumn{4}{c}{120B} & \multicolumn{4}{c}{180B} \\
\cmidrule(lr){2-5}\cmidrule(lr){6-9}\cmidrule(l){10-13}
Task & SPT & PPT & \shortstack{PPT\\$\to$SPT} & \shortstack{SPT\\$\to$PPT} & SPT & PPT & \shortstack{PPT\\$\to$SPT} & \shortstack{SPT\\$\to$PPT} & SPT & PPT & \shortstack{PPT\\$\to$SPT} & \shortstack{SPT\\$\to$PPT} \\
\midrule
\multicolumn{13}{@{}l}{English} \\
ARC-Challenge & 37.20 & 37.37 & 38.40 & 36.77 & 37.80 & 36.43 & 40.53 & 37.71 & 40.61 & 39.68 & 40.87 & 40.87 \\
HellaSwag & 50.05 & 48.77 & 51.46 & 51.37 & 51.61 & 50.90 & 53.62 & 53.62 & 53.21 & 50.88 & 55.56 & 55.66 \\
TruthfulQA & 22.64 & 23.38 & 23.62 & 22.52 & 22.15 & 24.60 & 23.13 & 22.28 & 23.99 & 24.72 & 22.03 & 23.26 \\
PAWS-X & 57.35 & 55.95 & 56.00 & 55.95 & 60.05 & 55.10 & 54.85 & 57.25 & 56.10 & 56.95 & 54.20 & 56.20 \\
XNLI & 47.31 & 47.27 & 46.79 & 47.71 & 49.48 & 47.71 & 46.35 & 47.75 & 47.43 & 46.99 & 46.83 & 46.99 \\
XWinograd & 72.73 & 73.89 & 74.11 & 74.71 & 74.80 & 73.42 & 76.00 & 75.40 & 74.19 & 76.52 & 77.16 & 77.03 \\
Mean & 47.88 & 47.77 & 48.40 & 48.17 & 49.32 & 48.03 & 49.08 & 49.00 & 49.26 & 49.29 & 49.44 & 50.00 \\
\addlinespace
\multicolumn{13}{@{}l}{French} \\
ARC-Challenge & 36.36 & 37.38 & 37.38 & 37.55 & 39.69 & 38.32 & 39.26 & 40.12 & 39.18 & 38.84 & 40.46 & 39.44 \\
HellaSwag & 45.12 & 45.79 & 46.76 & 47.13 & 46.94 & 46.59 & 48.53 & 48.50 & 48.33 & 47.01 & 49.67 & 50.03 \\
TruthfulQA & 25.16 & 25.92 & 24.40 & 27.70 & 26.43 & 26.18 & 24.65 & 26.18 & 28.34 & 27.19 & 26.43 & 27.32 \\
PAWS-X & 53.30 & 54.70 & 54.50 & 55.35 & 57.25 & 52.55 & 53.40 & 53.75 & 55.65 & 52.25 & 53.45 & 56.10 \\
XNLI & 44.26 & 43.41 & 44.18 & 45.34 & 45.18 & 42.17 & 43.25 & 44.62 & 43.21 & 43.13 & 43.57 & 44.78 \\
XWinograd & 59.04 & 55.42 & 63.86 & 60.24 & 61.45 & 60.24 & 69.88 & 63.86 & 56.63 & 67.47 & 61.45 & 61.45 \\
Mean & 43.87 & 43.77 & 45.18 & 45.55 & 46.16 & 44.34 & 46.50 & 46.17 & 45.22 & 45.98 & 45.84 & 46.52 \\
\bottomrule
\end{tabular*}
\captionof{table}{English--French task accuracy (\%). Mean averages six tasks per language.}\label{tab:tasks-enfr}

\par\vspace{\floatsep}
\setlength{\tabcolsep}{1.5pt}
\begin{tabular*}{\textwidth}{@{\extracolsep{\fill}}lrrrr@{\hspace{6pt}}rrrr@{\hspace{6pt}}rrrr@{}}
\toprule
 & \multicolumn{4}{c}{80B} & \multicolumn{4}{c}{120B} & \multicolumn{4}{c}{180B} \\
\cmidrule(lr){2-5}\cmidrule(lr){6-9}\cmidrule(l){10-13}
Condition & SPT & PPT & \shortstack{PPT\\$\to$SPT} & \shortstack{SPT\\$\to$PPT} & SPT & PPT & \shortstack{PPT\\$\to$SPT} & \shortstack{SPT\\$\to$PPT} & SPT & PPT & \shortstack{PPT\\$\to$SPT} & \shortstack{SPT\\$\to$PPT} \\
\midrule
\multicolumn{13}{@{}l}{EN$\to$FR} \\
FLORES A & 16.76 & 0.00 & 34.22 & 18.37 & 21.66 & 10.64 & 39.24 & 9.42 & 24.09 & 25.75 & 40.99 & 32.41 \\
FLORES B & 16.44 & 0.00 & 32.97 & 27.00 & 22.63 & 21.42 & 39.03 & 40.90 & 24.46 & 32.52 & 40.76 & 41.99 \\
WMT14 A & 13.89 & 0.00 & 26.20 & 12.64 & 15.37 & 5.44 & 30.56 & 6.21 & 17.25 & 12.75 & 32.76 & 22.66 \\
WMT14 B & 13.29 & 0.00 & 25.99 & 21.16 & 15.57 & 15.32 & 29.31 & 32.55 & 18.60 & 19.61 & 32.57 & 33.54 \\
\addlinespace
\multicolumn{13}{@{}l}{FR$\to$EN} \\
FLORES A & 22.04 & 0.46 & 35.80 & 14.02 & 21.32 & 11.34 & 37.13 & 18.18 & 25.09 & 25.97 & 37.96 & 37.81 \\
FLORES B & 24.15 & 0.00 & 28.39 & 35.43 & 21.43 & 21.06 & 36.70 & 37.66 & 27.18 & 17.16 & 37.07 & 40.61 \\
WMT14 A & 16.48 & 1.47 & 29.53 & 4.32 & 17.94 & 9.68 & 30.05 & 4.36 & 19.58 & 18.27 & 32.41 & 25.08 \\
WMT14 B & 19.24 & 0.00 & 22.60 & 28.51 & 18.07 & 18.90 & 30.22 & 29.45 & 22.07 & 13.40 & 31.60 & 32.46 \\
\addlinespace
Mean & 17.79 & 0.24 & 29.46 & 20.18 & 19.25 & 14.23 & 34.03 & 22.34 & 22.29 & 20.68 & 35.77 & 33.32 \\
\bottomrule
\end{tabular*}
\captionof{table}{English--French corpus BLEU at matched training budgets. FLORES denotes FLORES-200; A and B denote the prompts in Table~\ref{tab:translation-prompts}. Mean weights the eight conditions equally.}\label{tab:enfr-translation}

\par\endgroup
\end{minipage}
}]

\clearpage
\begin{table*}[!t]
\centering\small
\setlength{\tabcolsep}{1pt}
\begin{tabular*}{\linewidth}{@{\extracolsep{\fill}}lrrrrrr@{}}
\toprule
 & \multicolumn{2}{c}{100B starts} & \multicolumn{4}{c}{120B models} \\
\cmidrule(lr){2-3}\cmidrule(l){4-7}
Task & SPT & PPT & SPT & PPT & \shortstack{PPT\\$\to$SPT} & \shortstack{SPT\\$\to$PPT} \\
\midrule
\multicolumn{7}{@{}l}{English} \\
XNLI & 47.31 & 47.07 & 45.06 & 45.74 & 48.27 & 46.95 \\
XStoryCloze & 64.46 & 62.87 & 64.00 & 63.73 & 64.26 & 65.52 \\
COPA & 72.00 & 67.00 & 74.00 & 70.00 & 70.00 & 76.00 \\
Global-MMLU-Lite & 27.00 & 23.50 & 29.25 & 25.00 & 27.50 & 28.50 \\
Mean & 52.69 & 50.11 & 53.08 & 51.12 & 52.51 & 54.24 \\
\addlinespace
\multicolumn{7}{@{}l}{Swahili} \\
XNLI & 43.09 & 42.13 & 42.13 & 39.64 & 43.94 & 43.45 \\
XStoryCloze & 56.12 & 55.92 & 56.06 & 55.72 & 56.52 & 56.32 \\
XCOPA & 59.00 & 60.20 & 58.80 & 59.20 & 62.40 & 59.40 \\
Global-MMLU-Lite & 25.50 & 24.25 & 25.75 & 25.50 & 28.00 & 24.75 \\
Mean & 45.93 & 45.63 & 45.69 & 45.02 & 47.72 & 45.98 \\
\bottomrule
\end{tabular*}
\caption{English--Swahili task accuracy (\%) before (100B) and after (120B) continuation. Mean averages four tasks per language.}\label{tab:ensw-tasks}
\par\vspace{\floatsep}
\noindent\begin{minipage}{\columnwidth}
\centering\small
\setlength{\tabcolsep}{.5pt}
\begin{tabular*}{\linewidth}{@{\extracolsep{\fill}}lrrrr@{}}
\toprule
Condition & SPT & PPT & \shortstack{PPT\\$\to$SPT} & \shortstack{SPT\\$\to$PPT} \\
\midrule
EN$\to$SW A & 16.37 & 13.56 & 25.37 & 0.45 \\
EN$\to$SW B & 14.60 & 22.91 & 22.11 & 2.15 \\
SW$\to$EN A & 2.44 & 0.72 & 2.00 & 1.08 \\
SW$\to$EN B & 2.86 & 11.94 & 3.22 & 2.50 \\
\midrule
Mean & 9.07 & 12.28 & 13.18 & 1.54 \\
\bottomrule
\end{tabular*}
\captionof{table}{English--Swahili corpus BLEU at 120B on 1,012 FLORES-200 sentences. A and B denote the prompts in Table~\ref{tab:translation-prompts}. Mean weights both directions and prompts equally.}\label{tab:ensw-translation}
\end{minipage}\hfill\mbox{}
\end{table*}

\end{document}